\documentclass{article}

\usepackage{PRIMEarxiv}

\usepackage[utf8]{inputenc} 
\usepackage[T1]{fontenc}    
\usepackage{hyperref}       
\usepackage{url}            
\usepackage{booktabs}       
\usepackage{amsfonts}       
\usepackage{nicefrac}       
\usepackage{microtype}      
\usepackage{lipsum}
\usepackage{graphicx}
\graphicspath{{media/}}     

\title{Approach to Designing CV Systems for Medical Applications: Data, Architecture and AI
}

\author{
  Dmitry Ryabtsev \\
  HSE University \\
  Moscow, Russia \\
  \texttt{diryabtsev@edu.hse.ru} \\
   \And
  Boris Vasilyev \\
  Utrecht University \\
  Utrecht, Netherlands \\
  \texttt{b.a.vasilyev@students.uu.nl} \\
   \And
  Sergey Shershakov \\
  HSE University \\
  Moscow, Russia \\
  \texttt{sshershakov@hse.ru} \\ 
}

\begin{document}
\maketitle

\begin{abstract}
This paper introduces an innovative software system for fundus image analysis that deliberately diverges from the conventional screening approach, opting not to predict specific diagnoses. Instead, our methodology mimics the diagnostic process by thoroughly analyzing both normal and pathological features of fundus structures, leaving the ultimate decision-making authority in the hands of healthcare professionals~\cite{topol2019highperformance}.

Our initiative addresses the need for objective clinical analysis and seeks to automate and enhance the clinical workflow of fundus image examination. The system, from its overarching architecture to the modular analysis design powered by artificial intelligence (AI) models, aligns seamlessly with ophthalmological practices. Our unique approach utilizes a combination of state-of-the-art deep learning methods and traditional computer vision algorithms to provide a comprehensive and nuanced analysis of fundus structures.

We present a distinctive methodology for designing medical applications, using our system as an illustrative example. Comprehensive verification and validation results demonstrate the efficacy of our approach in revolutionizing fundus image analysis, with potential applications across various medical domains.
\end{abstract}


\section{Introduction}
Artificial intelligence (AI) has demonstrated immense potential in revolutionizing medical imaging diagnostics~\cite{lundervold2019overview,litjens2017survey}, particularly through advancements in computer vision (CV). The ability of AI to analyze complex patterns in medical images—such as MRIs, X-rays, CT scans, and photographs—has led to significant progress in disease detection and classification. However, despite these technological advancements, there remains a substantial disconnect between the design of AI systems and the practical needs of medical professionals. The relentless race for higher performance often overlooks critical factors such as interpretability, transparency~\cite{rudin2019interpretable,holzinger2019explainable}, and integration into clinical workflows, which are essential for real-world adoption in healthcare settings.

One of the major challenges for AI in medicine is data scarcity. Medical data is not only limited in quantity but often lacks the quality and relevance necessary for effective AI model training~\cite{kaushal2020bias,cheplygina2019scarcity}. The absence of standardized protocols for data anonymization, along with domain-specific data difficulties and interobserver variability, further complicates data collection and curation. Additionally, the current trend toward large, complex models amplifies issues of interpretability and transparency, as these models often function as "black boxes," making it difficult for clinicians to understand and trust their outputs.

Aligning AI systems with clinical workflows is crucial to bridge this gap. In medical imaging, and particularly in ophthalmology, there is a significant opportunity to leverage CV for fundus image analysis. Fundus images provide valuable insights into ocular and systemic diseases, and effective analysis tools can greatly assist clinicians. However, for AI to be truly effective in this domain, it must be designed with the real needs of doctors in mind.

In this paper, we introduce EYAS, an AI-powered fundus image analysis system that exemplifies our clinician-centric design approach. EYAS addresses the challenges of data scarcity through careful data curation involving medical specialists at every stage, ensuring that the data is both high-quality and clinically relevant. The system is designed to be modular and adaptable, incorporating specialized modules for specific tasks and generating intermediate results that are understandable and verifiable by clinicians. By prioritizing transparency and aligning closely with clinical workflows, EYAS aims to bridge the gap between cutting-edge AI technology and the essential requirements of healthcare professionals, facilitating the meaningful adoption of AI in ophthalmology and beyond.

\section{Related Work}

Screening systems have become prevalent in fundus image analysis~\cite{gulshan2016development} due to opportunistic data acquisition, enabling binary classifications such as the presence or absence of disease \cite{ting2019artificial}. While these systems leverage large datasets that are more readily available, they often lack the detailed annotations necessary for nuanced analyses. This approach faces significant limitations related to data scarcity, lack of standardization, and insufficient interpretability.

Data scarcity is a critical issue in medical imaging AI. The absence of universal standards for anonymizing medical images complicates data sharing and aggregation, hindering the accumulation of large, diverse datasets required for robust AI model development \cite{ting2019artificial}. Additionally, there is a paucity of research on demographic differences, which is vital for ensuring that AI models perform effectively across diverse populations. Domain-specific challenges, such as variations in imaging equipment and protocols, further complicate data collection and model training \cite{holzinger2020artificial}. Interobserver variability—the differences in interpretation among clinicians—impacts the consistency and reliability of annotated data, affecting model performance \cite{hicks2022on}.

Since data is the foundation of any AI system, the importance of data quality and curation cannot be overstated. Poor-quality data is not merely less useful; it is often not useful at all \cite{albarqouni2016aggnet}. Effective curation requires collaboration between system designers and clinicians to ensure that both the output and intermediate data meet clinical needs. If custom outputs are necessary, systems must be designed to produce them, emphasizing the need for flexibility in AI system design.

Transparency and interpretability are crucial for ensuring reliability and safety in medical AI applications. Designing evaluation metrics and validation strategies that the system can provide is essential for building trust among clinicians \cite{band2023application}. Intermediate data must be understandable to clinicians so they can discern which predictions are based on specific inputs, facilitating verification and fostering confidence in the system \cite{sarao2023explainable}.

Segmentation-based approaches offer an alternative to screening systems by providing detailed analyses of specific structures within fundus images \cite{ronneberger2015u}. These methods enhance interpretability and allow for outputs that can be directly integrated into clinical reports, aligning more closely with the diagnostic processes used by ophthalmologists. However, challenges in acquiring finely annotated data have limited the widespread adoption of segmentation approaches.

A modular approach that mirrors clinical workflows remains a largely unexplored avenue in fundus image analysis. Such an approach allows for step-by-step verification, harnessing the potential of explainable AI and offering a compelling path toward enhanced clinical trust in AI decision-making \cite{liu2019self}.

\section{Requirements Analysis}
\begin{figure*}[!h]
\centering
\includegraphics[width=0.8\textwidth]{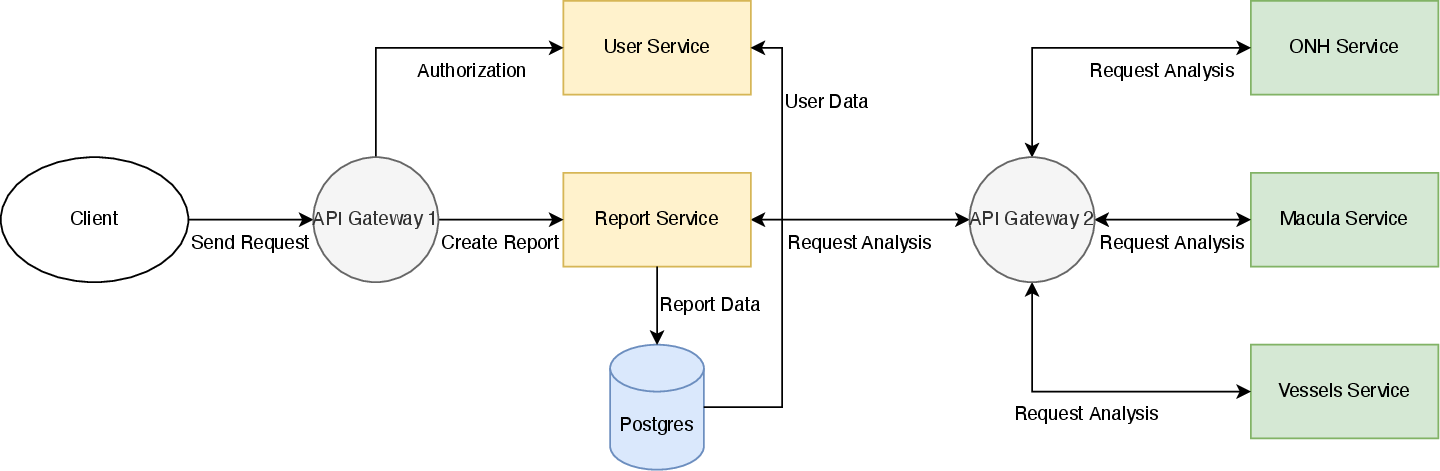}
\caption{Architecture of the EYAS fundus image analysis system, illustrating the client-server model, API gateways, analysis microservices for fundus structures, and the report service that synthesizes results for clinical use.}
\label{fig1}
\end{figure*}

\textbf{Scope of Automation.}
The first step in the development of an AI-driven system for ophthalmology is to carefully delineate the aspects of the clinical workflow that can be effectively automated. We focus on analyzing fundus images and generating detailed, natural language descriptions of fundus structures. This capability allows the system to produce a draft of a medical report that closely aligns with the format used by ophthalmologists. By automating this specific component of the clinical workflow, the system aims to significantly enhance operational efficiency. The scope of automation is deliberately narrowed to ensure reliability and to avoid overextending the system's capabilities. This approach ensures that the system provides meaningful support to clinicians, helping them to streamline their documentation tasks without undermining the critical decision-making processes inherent in medical practice.

\textbf{Data Quality and Curation.}
Data quality and curation are foundational to the success of any AI-based medical system. It is essential that the system is developed with data that meets the expectations and needs of its primary users—clinicians. To achieve this, we prioritize engaging with end users during the data curation process to ensure that the collected data is both relevant and user-friendly in a clinical context. The data collection process should adhere to established medical protocols, which are the closest available standards for ensuring data quality in medical applications. This process includes addressing domain-specific challenges such as class imbalances, where pathological cases may be underrepresented in the dataset. Properly curated and balanced data is critical for the system to deliver reliable and clinically useful outputs.

\textbf{System Efficiency and Feasibility.}
In addition to ensuring a clear scope and high data quality, it is crucial to assess the feasibility and efficiency of building and deploying the system. This involves determining whether the proposed system can be constructed using current AI technologies in a manner that is time-efficient, cost-effective, and integratable within existing clinical workflows. The system must offer practical benefits, such as saving clinicians time, reducing costs, and providing rapid analysis, to justify its development and implementation. Furthermore, the system's design should facilitate seamless integration into existing medical infrastructures, minimizing disruption and ensuring that it complements rather than complicates the clinician's workflow.

\section{Data Challenges and Strategies}
\begin{figure*}[!h]
\centering
\includegraphics[width=0.8\textwidth]{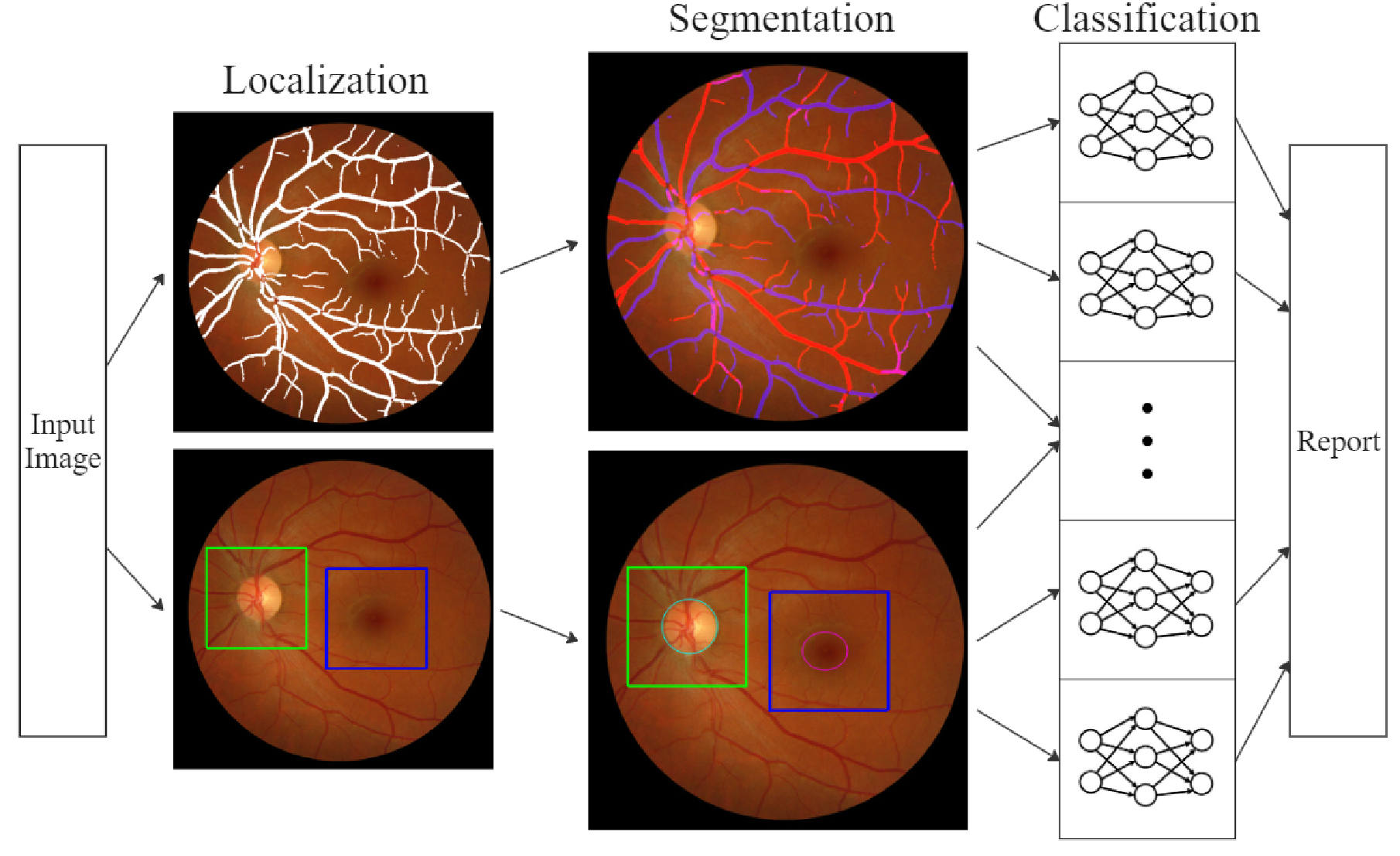} 
\caption{The modular analysis pipeline of EYAS, showing the three-step process of localization, segmentation, and classification for fundus characterization.}
\label{fig2}
\end{figure*}

The successful implementation of AI systems in medical imaging is heavily dependent on the quality of the annotated data used for training and validation. Only domain experts—medical specialists with extensive knowledge and experience—are viable annotators for such data \cite{holzinger2020artificial}. Their involvement is crucial to ensure that annotations are accurate, clinically relevant, and encompass the specific features that are essential for diagnosis and treatment planning.

Medical specialists must be actively involved in every stage of data collection and annotation. Focusing on specific features over general ones ensures that the data aligns with clinical reporting practices \cite{albarqouni2016aggnet}. This specificity enhances the utility of the AI system in real-world clinical settings, as the outputs are directly translatable into actionable insights for patient care.

In addition to initial data annotation, medical professionals should be involved in designing and verifying intermediate data outputs. These outputs must be sufficient for diagnostic purposes and comprehensible enough for clinicians to verify without compromising necessary details \cite{sarao2023explainable}. There can be no compromise on the necessity of this data in the system's design; it must meet the exacting standards of clinical practice to be trustworthy and effective.

Medical professionals also play a critical role in informing system designers about any present imbalances or specific difficulties within the data. Issues such as class imbalances—where certain pathologies may be underrepresented—can significantly affect model performance \cite{hensman2013gaussian}. Their insights are essential in guiding the data curation process to ensure that the dataset is both comprehensive and representative of the patient populations the AI system will serve.

Collaborative involvement extends to the design of evaluation metrics and validation strategies. Traditional performance metrics may not adequately capture the nuances of clinical significance \cite{hicks2022on}. Medical experts can help define metrics that reflect the real-world importance of different types of errors, leading to more clinically relevant evaluations of model performance. This collaboration also influences the selection of loss functions during model training, aligning the optimization process with clinical priorities \cite{salehi2017tversky}.

Addressing data challenges requires proactive strategies:

\begin{itemize} \item \textbf{Targeted Data Collection:} Actively seeking out data for underrepresented classes to balance the dataset \cite{ting2019artificial}. \item \textbf{Data Augmentation:} Using techniques such as rotation, scaling, and flipping to increase data diversity and mitigate overfitting \cite{ronneberger2015u}. \item \textbf{Advanced Annotation Techniques:} Employing methods like consensus labeling to reduce interobserver variability and improve annotation quality \cite{albarqouni2016aggnet}. \end{itemize}

By closely involving medical professionals in these processes, the AI system is more likely to meet the stringent requirements of clinical application, ultimately leading to safer and more effective patient care. The collaboration ensures that the system not only performs well statistically but also provides outputs that are meaningful and trustworthy from a clinical perspective.

\section{General Methodology and Approach}
Our approach to developing the fundus image analysis system is a direct outcome of the requirements analysis, which identified the key areas where automation can be effective, data quality must be ensured, and system efficiency is paramount. The methodology outlined below reflects these priorities, emphasizing modular design, appropriate technology selection, interpretability, and seamless integration into both diagnostic and clinical workflows.

\textbf{Modular Approach.}
The development of our system follows a modular approach, designed to enhance flexibility, scalability, and clarity. By breaking down the image analysis process into discrete tasks—such as localization, segmentation, and classification—each module can be optimized individually, ensuring that the overall system remains efficient and effective. This modular design not only improves the interpretability of the system's outputs but also allows for easier updates and maintenance, enabling the system to adapt to evolving clinical needs and technological advancements.

\textbf{Appropriate Technology Selection.}
In line with our commitment to efficiency and practicality, we carefully select the most suitable technology for each task. This means using traditional computer vision algorithms when they are sufficient and opting for deep learning models only when the complexity of the task necessitates it. This approach avoids unnecessary complexity, ensuring that the system is not only effective but also accessible and interpretable for clinicians. By focusing on the right tool for the job, we strike a balance between performance and usability, ensuring that the system remains a practical and reliable tool in clinical settings.

\textbf{Emphasis on Interpretability.}
Interpretability is a cornerstone of our system's design, as it is crucial for gaining the trust of clinicians who will rely on the system in their daily practice. The system is designed to produce outputs that are easily interpretable by doctors, allowing them to quickly verify the results. In addition to the final outcomes, the system provides transparency in its intermediate results, offering insights into how conclusions are reached at each stage of the analysis. This transparency helps clinicians understand and trust the process, ensuring that they can confidently integrate the system into their diagnostic workflow, using it as a reliable aid rather than a black-box tool.

\textbf{Diagnostic Workflow Integration.}
Our system is meticulously designed to integrate seamlessly with the diagnostic workflow of ophthalmologists. It mirrors the way doctors describe and analyze fundus images in practice, generating detailed descriptions that align with clinical protocols and practices. This approach ensures that the system's outputs are directly relevant and useful in a clinical context, providing doctors with actionable insights that are presented in a familiar format. The system's design prioritizes practical utility, making it a natural extension of the clinician's existing diagnostic processes.

\textbf{Clinical Workflow Integration.}
Beyond diagnostic tasks, the system is engineered to fit smoothly into the broader clinical workflow. It is fully compatible with existing Electronic Medical Records (EMR) systems, ensuring that data can be easily integrated and accessed within the established healthcare infrastructure. The system also adheres to stringent security and anonymity standards, safeguarding patient data and ensuring compliance with healthcare regulations. By being EMR-ready and prioritizing security, the system supports a seamless and secure integration into the clinical environment, enabling its use across diverse healthcare settings without disrupting existing operations.

\section{Details of Implementation}

\subsection{Architecture Design}

\begin{figure*}[!h]
\centering
\includegraphics[width=0.8\textwidth]{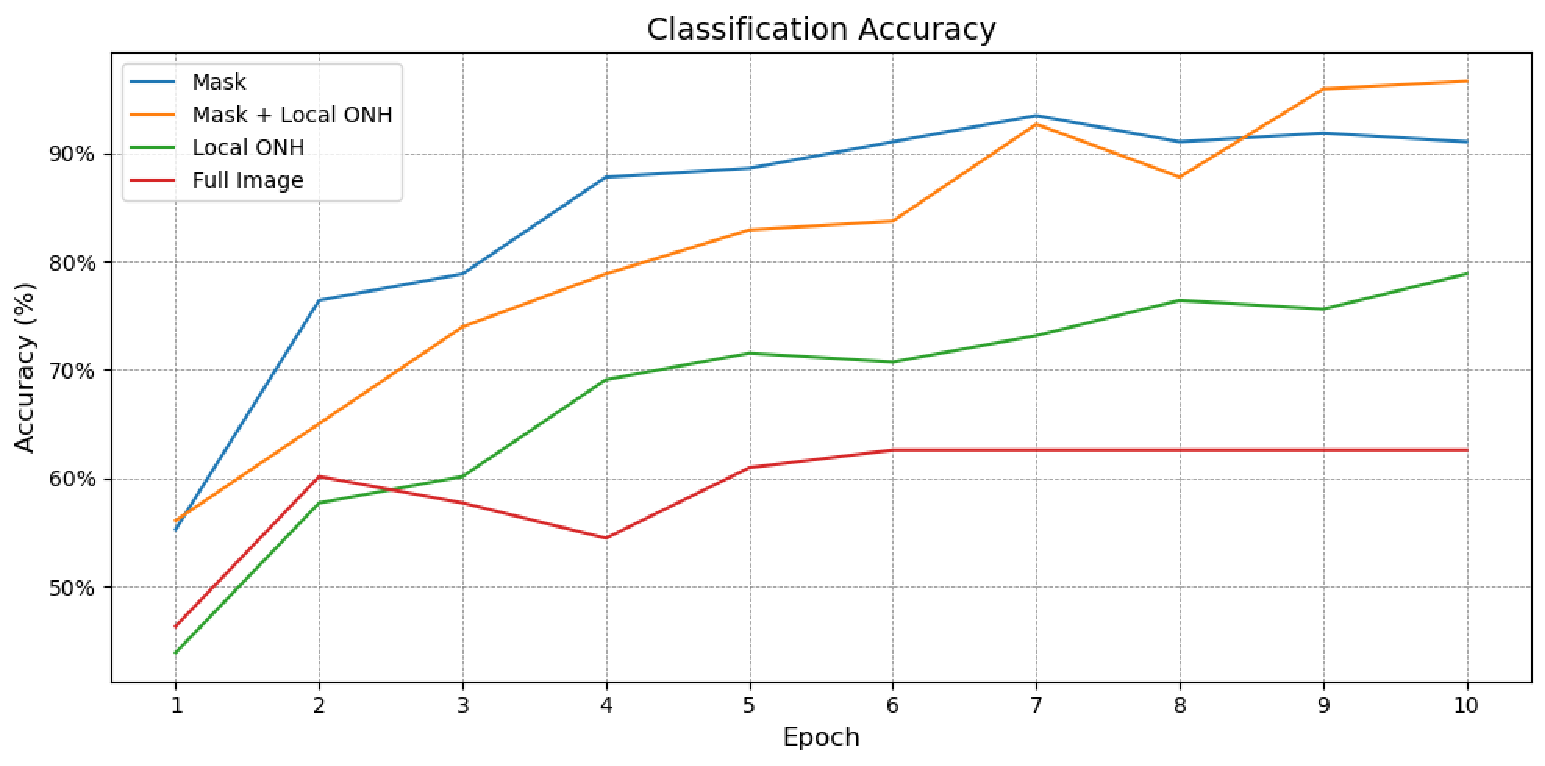}
\caption{Performance comparison of \texttt{ONH Shape} classifiers using different input formats, highlighting the impact of the analysis pipeline steps on classification accuracy.} \label{fig3}
\end{figure*}

The architecture of our fundus image analysis system (see Fig.~\ref{fig1}) is carefully designed to ensure modularity, scalability, and seamless integration into clinical workflows. It follows a client-server model where the interaction between the web client and the backend services is managed through API Gateways. The first gateway handles the client interface, while the second is dedicated to optimizing the operations of the analysis microservices, which are responsible for specific aspects of fundus image analysis.

At the core of the system is the Report Service, which synthesizes analysis results from various microservices, each focused on distinct fundus structures such as the optic nerve head (ONH), macula, and retinal vessels. These microservices operate autonomously but are designed to communicate with each other, enabling cross-analysis. This means that the results of one parameter analysis can inform and enhance another, even when these analyses are conducted by different microservices. For example, the analysis of vessel characteristics can be influenced by findings from the macula or ONH analysis, ensuring a more comprehensive and integrated assessment of the fundus.

Our architecture is closely aligned with the clinical decision-making process. Each microservice corresponds directly to the specific diagnostic decisions and descriptive elements that clinicians consider during fundus image assessment. The ONH microservice, for instance, focuses on features crucial for diagnosing conditions like glaucoma, while the macula and vessels microservices are tailored to their respective clinical tasks. This alignment ensures that the system's outputs are not only technically accurate but also clinically relevant, providing doctors with detailed and actionable insights that support their diagnostic process.

To accommodate the rapid pace of advancements in AI and model development, the system is designed with a plug-and-play approach. Each microservice can be updated or replaced independently, allowing for the seamless integration of new models or improvements without disrupting the overall system. This flexibility ensures that the system remains at the forefront of technological innovation, adapting quickly to changes in clinical practice and maintaining its effectiveness in a rapidly evolving field.

Overall, the architecture is built to provide a robust and adaptable solution that not only meets the immediate needs of clinicians but also supports ongoing innovation and improvement in the analysis of fundus images. The system is designed to complement the analysis workflow that clinicians already use, ensuring that it integrates smoothly into their existing practices while enhancing their ability to make accurate and informed diagnostic decisions.

\subsection{Analysis Pipeline}

Our AI-based fundus analysis system employs a three-step modular pipeline (see Fig.~\ref{fig2}) designed to enhance interpretability, promote model explainability, and optimize performance.

\textbf{Localization} We utilize an ensemble of OpenCV~\cite{opencv_library} methods for precise localization of the ONH and macula regions within the original ophthalmoscopy image. This includes image preprocessing for contrast enhancement, edge detection, and template matching for robust detection.

\textbf{Segmentation} For ONH and Macula we employ state-of-the-art U-Net variants (e.g., Attention U-Net, R2U-Net ~\cite{Oktay2018-ag,Alom2018-qk}) trained on a carefully curated subset of the public RIGA dataset, ensuring a representative training sample.  Vessel system segmentation is performed by a pre-trained W-Net model~\cite{Galdran2020-su} trained on HRF~\cite{Budai2013-fx} and DRIVE~\cite{Drive} datasets, with a two-stage process for vessel identification and subsequent artery/vein differentiation.

\textbf{Classification} We utilize various classification models, including transfer learning of ResNet and DenseNet ~\cite{He2015-ih,Huang2016-rv} architectures, showcasing our system's flexibility. These models analyze the precise masks generated during segmentation to classify an exhaustive range of clinically relevant characteristics such as optic disk shape, artery caliber, macular reflex.

This modular approach strategically decomposes fundus analysis. Initial focus on localization improves interpretability and downstream model performance. Advanced U-Net variants enhance segmentation accuracy for precise analysis of subtle characteristics. Our flexible classification step enables adaptation to diverse diagnostic tasks, aligning with key aspects of the clinical workflow to facilitate medical expert verification and promote trust in the AI’s outputs.

\section{Verification and Validation}

To ensure the efficacy and reliability of our proposed analysis pipeline, we implemented comprehensive validation and verification tailored to each stage of our three-step processing.

\textbf{Validation} During training, each model's performance was monitored using established machine learning metrics~\cite{Hicks2022-rp} carefully selected for their relevance to each step. For segmentation tasks, we prioritized precision and Intersection over Union (IoU) due to the small proportional area of segmentation regions. In classification tasks, we utilize standard accuracy, along with per-class accuracy analysis to ensure the model performs well for both normal and pathological cases.

\textbf{Verification} Validation alone is not enough to guarantee seamless integration of AI into clinical practice. We complemented our quantitative validation with a verification step focused on ensuring that the outputs align with clinical expectations and workflows. Medical professionals provided detailed feedback on the system's outputs, which informed iterative refinements to the models and user interface. Their insights ensured that the system's results were clinically relevant and aligned with diagnostic expectations, enhancing both accuracy and usability.

\subsection{Performance evaluation}

Figure~\ref{fig3} exemplifies the impact of our pipeline on classification accuracy. Here, we compare four ResNet-based models, all with similar architectures but adapted for different input formats. Validation was performed on a 20\% split of our ONH dataset.

\textbf{Image and Local ONH} These models take the full fundus image and the localized optic disk image, respectively. Their sub-optimal performance suggests that relying solely on raw image information may not be sufficient for optimal classification.

\textbf{Mask and Mask + Local ONH} These models utilize a more informative input format. They incorporate the mask and Local ONH with the mask added as an additional channel. The superior performance of these models demonstrates that the feature extraction achieved through our analysis pipeline not only enhances explainability but also leads to stronger predictive power.

\section{Conclusion}

The development and implementation of the EYAS fundus image analysis system exemplify our innovative, clinician-centric approach to designing medical applications. By aligning closely with ophthalmological workflows and prioritizing interpretability and data quality, our system addresses critical needs in automated fundus image analysis.

Our modular architecture, leveraging both deep learning and traditional computer vision techniques, demonstrates flexibility, scalability, and seamless integration into existing clinical practices. Positive evaluations by medical experts validate not only the technical efficacy of our system but also its clinical utility, underscoring the importance of collaboration between technologists and healthcare professionals.

By providing detailed fundus structure descriptions rather than relying on traditional screening approaches, we empower clinicians with comprehensive, objective insights that enhance diagnostic accuracy. Our approach bridges the gap between advanced AI technologies and clinical expertise, setting the stage for future innovations in medical AI applications.

Looking ahead, our methodology offers a blueprint for developing AI systems in other medical domains. Emphasizing collaboration, interpretability, and workflow integration, we aim to facilitate meaningful adoption of AI in healthcare, ultimately improving patient care across various specialties.

\bibliographystyle{unsrt}  
\bibliography{references}

\end{document}